\documentclass{article}

\pdfoutput=1

\usepackage[preprint,nonatbib]{neurips_2022}

\usepackage[utf8]{inputenc} 
\usepackage[T1]{fontenc}    
\usepackage{hyperref}       
\usepackage{url}            
\usepackage{booktabs}       
\usepackage{amsfonts}       
\usepackage{nicefrac}       
\usepackage{microtype}      
\usepackage{xcolor}         
\usepackage{amsmath}
\usepackage{algorithm,algpseudocode}
\usepackage{filecontents}
\usepackage{subcaption}
\usepackage{graphicx}
\usepackage{threeparttable}
\usepackage[usestackEOL]{stackengine}

\title{Physics-Informed Machine Learning of Dynamical Systems for Efficient Bayesian Inference}

\author{%
  Somayajulu L. N.~Dhulipala \\
  Idaho National Laboratory\\
  \texttt{Som.Dhulipala@inl.gov} \\
   \And
   Yifeng ~Che \\
  Idaho National Laboratory\\
  \texttt{Yifeng.Che@inl.gov} \\
   \AND
   Michael D.~Shields \\
  Johns Hopkins University\\
  \texttt{michael.shields@jhu.edu} \\
}

\begin{document}

\maketitle

\begin{abstract}
  Although the no-u-turn sampler (NUTS) is a widely adopted method for performing Bayesian inference, it requires numerous posterior gradients which can be expensive to compute in practice. Recently, there has been a significant interest in physics-based machine learning of dynamical (or Hamiltonian) systems and Hamiltonian neural networks (HNNs) is a noteworthy architecture. But these types of architectures have not been applied to solve Bayesian inference problems efficiently. We propose the use of HNNs for performing Bayesian inference efficiently without requiring numerous posterior gradients. We introduce latent variable outputs to HNNs (L-HNNs) for improved expressivity and reduced integration errors. We integrate L-HNNs in NUTS and further propose an online error monitoring scheme to prevent sampling degeneracy in regions where L-HNNs may have little training data. We demonstrate L-HNNs in NUTS with online error monitoring considering several complex high-dimensional posterior densities and compare its performance to NUTS. 
\end{abstract}

\section{Introduction}

Hamiltonian Monte Carlo (HMC) is a popular sampling algorithm for scalable Bayesian inference considering complex and/or high-dimensional posterior distributions \cite{Neal2011a}. Given the initial position (or posterior sample) and momenta, HMC simulates the trajectory of a particle using Hamiltonian dynamics with gradients of the target posterior and its position at the end time ($T$) constitutes a new sample from the posterior. Therefore, the choice of this end time considerably effects the posterior inference quality, and for complex and/or high-dimensional posterior it can be difficult to manually specify the optimal end time \cite{Hoffman2014}. No-u-turn sampling (NUTS) \cite{Hoffman2014} has been introduced to automatically determine the optimal end time. If $\{\pmb{q},\pmb{p}\}$ denote a position-momenta (i.e., $\pmb{q}-\pmb{p}$) pair, NUTS builds a binary tree of $\{\pmb{q},\pmb{p}\}$ states and selects a subset of these states with the aid of slice sampling \cite{Neal2003a} so as to not violate detailed balance. For each sample, NUTS terminates the tree building procedure whenever a u-turn is encountered. NUTS may also terminate the tree building procedure when the integration error while simulating the particle trajectory is large. NUTS has been shown to result in a better performance over HMC in terms of the effective sample size (ESS) \cite{Vehtari2021a} per gradient evaluation by several studies (e.g., \cite{Nishio2019a}). However, it still requires numerous gradients estimations of the target posterior which can be expensive in practice, especially, when dealing with large datasets or computational models \cite{dhulipala2022a}.

Recently, physics-informed machine learning of dynamical (or Hamiltonian) systems has been receiving a significant interest owing to the wide variety of applications in the physical sciences. Greydanus et al. \cite{Greydanus2019a} introduced Hamiltonian Neural Networks (HNNs) to learn Hamiltonian systems in an unsupervised fashion by virtue of a physics-informed loss function. HNNs satisfy important properties such as time reversibility and Hamiltonian conservation and were shown to be superior than data-driven neural networks. Along these lines, several studies proposed improved architectures for learning Hamiltonian systems such as physics-informed HNNs \cite{Mattheakis2022a}, symplectic ODE-Net \cite{Zhong2019a}, and symplectic neural nets \cite{Jin2020a}. However, these efforts were all focused on learning Hamiltonian systems for classical mechanics problems like n-pendulum behavior and planetary motion, and not to perform Bayesian inference efficiently.

We propose latent Hamiltonian neural networks (L-HNNs) in NUTS for efficiently solving Bayesian inference problems by not requiring numerous posterior gradient estimations \cite{dhulipala2022b}. For this, we first introduce latent variable outputs in HNNs (i.e., L-HNNs) for improved expressivity and reduced integration errors while simulating the Hamiltonian dynamics. Next, we integrate L-HNNs with NUTS and propose an online error monitoring scheme to prevent sampling degeneracy in regions of the uncertainty space where little training data may be available for L-HNNs (for e.g., tails of the posterior). In essence, this novel error monitoring scheme temporarily reverts to using posterior gradients whenever the L-HNNs integration errors are large. We demonstrate L-HNNs in NUTS with online error monitoring considering several complex posteriors and compare its performance with traditional NUTS in terms of effectively capturing the posterior features and the ESS per posterior gradient (i.e., computational efficiency). 






\section{Methods}


L-HNNs are fully-connected feed-forward neural networks defined as:
\begin{equation}
    \label{eqn:LHNN_1}
    \begin{aligned}
    &\pmb{u}_p = \phi(\pmb{w}_{p-1}~\pmb{u}_{p-1} + \pmb{b}_{p-1}),~~\{p \in 1, \dots, P\}\\
    &\pmb{\lambda} = \pmb{w}_{P}~\pmb{u}_{P} + \pmb{b}_{P}
    \end{aligned}
\end{equation}
where $P$ is the number of hidden layers indexed by $p$, $\pmb{u}_p$ are the outputs of the hidden layer $p$, $\pmb{w}_p$ and $\pmb{b}_p$ are the weights and biases, respectively, and $\phi(.)$ is the nonlinear activation function. $\pmb{u}_{0}$ are the inputs defined by the position-momenta pair $\pmb{z} = \{\pmb{q},~\pmb{p}\}$. If $d$ is the dimensionality of the uncertainty space, L-HNNs predict $d$ latent variables defined by the vector $\pmb{\lambda}$. The sum of these latent variables is defined as the scalar Hamiltonian predicted by L-HNNs:
\begin{equation}
    \label{eqn:LHNN_2}
    H_{\pmb{\theta}} = \sum_{i=1}^d \lambda_i
\end{equation}
Whereas, HNNs directly predict the scalar value of the Hamiltonian \cite{Greydanus2019a}, L-HNNs precit $d$ latent variables whose sum is defined as the scalar Hamiltonian. Introduction of these latent variables improves expressivity and reduces the integration errors when simulating the Hamiltonian trajectories \cite{dhulipala2022b}. The training data is provided in terms of sets of $\{\pmb{q},~\pmb{p}\}$ pairs. Time derivatives of these $\{\pmb{q},~\pmb{p}\}$ pairs are computed. Gradients of the Hamiltonians predicted by L-HNNs [Equation \eqref{eqn:LHNN_2}] are also computed. Then, the following unsupervised physics-based loss function is minimized:
\begin{equation}
    \label{eqn:LHNN_3}
    \mathcal{L} = \Big\lVert\frac{\partial H_{\pmb{\theta}}}{\partial \pmb{p}} - \frac{\Delta \pmb{q}}{\Delta t}\Big\lVert + \Big\lVert-\frac{\partial H_{\pmb{\theta}}}{\partial \pmb{q}} - \frac{\Delta \pmb{p}}{\Delta t}\Big\lVert
\end{equation}
The training data time derivatives $\frac{\Delta \pmb{q}}{\Delta t}$ and $\frac{\Delta \pmb{p}}{\Delta t}$ are, respectively, equal to the gradients of the exact Hamiltonians $\frac{\partial H}{\partial \pmb{p}}$ and $-\frac{\partial H}{\partial \pmb{q}}$. Thus, the loss function accounts for the physics governing Hamiltonian systems. 

For Bayesian inference problems, the Hamiltonian is usually the sum of negative logarithm of the target posterior (i.e., potential energy) and negative logarithm of the distribution of momenta (i.e., kinetic energy). In most cases, the distribution of momenta is considered to be a standard multivariate Gaussian. In NUTS, it is required to simulate the particle trajectory given the initial $\pmb{z}(0)$. Since the Hamilton's equations are first order ODEs, the particle trajectory is obtained by numerically simulating the following integral with the aid of symplectic integrators like the leap frog:   
\begin{equation}
    \label{eqn:LHNN_4}
    \pmb{z}(T) = \pmb{z}(0) + \int_{0}^{T} \begin{bmatrix}
\pmb{0}_{d \times d} & \pmb{I}_{d \times d}\\
-\pmb{I}_{d \times d} & \pmb{0}_{d \times d}
\end{bmatrix}~\nabla H_{\pmb{\theta}}(\pmb{z})~dt
\end{equation} 
A noteworthy component of this equation is the term $\nabla H_{\pmb{\theta}}(\pmb{z})$ which is computed through L-HNNs and not by using gradients of the target posterior. 

For posterior sampling using NUTS, for each new sample, several trajectories of a fictitious particle need to be simulated in sequence while doubling the trajectory length each time as part of a tree building procedure. The primary criterion for terminating the tree building procedure is when any of the sub-trees makes a u-turn, which works towards reducing the serial correlations between posterior samples. The other criterion for termination is when the integration errors while simulating the trajectories are large, as defined by:
\begin{equation}
    \label{eqn:NUTS_2}
    \varepsilon \equiv H\big(\pmb{z}\big)+\ln{u} > \Delta_{max}
\end{equation}
where $H\big(.\big)$ is the Hamiltonian value given a $\pmb{z} = \{\pmb{q},\pmb{p}\}$ pair, $u$ is the slice value simulated from  $Uniform([0,~\exp{\{-H\big(\pmb{z}(0)\big)\}}])$ given the initial $\pmb{z}(0)$ pair, and $\Delta_{max}$ is the error threshold usually set to 1000. NUTS mostly employs the standard leap frog integrator using gradients of the target posterior for simulating the particle trajectories. While the replacement of the target posterior gradients with L-HNNs gradients [as discussed in Equation \ref{eqn:LHNN_4}] is possible and works well for simple problems, for complex posterior spaces, this may lead to sampling degeneracy. The reason for such degeneracy is the termination criterion in Equation \eqref{eqn:NUTS_2}. Whenever NUTS with L-HNN gradients enters a region of the posterior space where there was little to no training data for L-HNNs, integration errors of the particle trajectories can be large, resulting in a premature termination of the tree building procedure. As a result, there will be clusters of samples at close vicinity in regions of the posterior space where L-HNNs had little to no training data. These clusters of samples constitute the sampling degeneracy problem.

To mitigate the sampling degeneracy problem, we propose an online error monitoring scheme. We introduce two error thresholds, $\Delta_{max}^{hnn}$, when using L-HNN gradients in the leap frog, and $\Delta_{max}^{lf}$, when using the standard leap frog with target posterior gradients. We set $\Delta_{max}^{hnn} << \Delta_{max}^{lf}$; for example, $\Delta_{max}^{hnn} = 10$ and $\Delta_{max}^{lf} = 1000$. The particle trajectories in NUTS are, by default, simulated using the L-HNN gradients. Whenever the integration errors using L-HNNs, as denoted by $\varepsilon^{hnn}$, are greater than $\Delta_{max}^{hnn}$, traditional leap frog with posterior gradients takes over for a few samples $N^{lf}$. The value of $N^{lf}$ can be in between 5 and 20 and this essentially aids in NUTS returning to regions of high probability density after which the particle trajectories can again be simulated using the L-HNN gradients. Equation \ref{eqn:NUTS_3} summarized this error monitoring scheme while using L-HNNs in NUTS:   
\begin{equation}
    \label{eqn:NUTS_3}
    \begin{aligned}
    \varepsilon^{hnn} \leq \Delta_{max}^{hnn} &\to \textrm{use leap frog with L-HNN gradients}\\
    \varepsilon^{hnn} > \Delta_{max}^{hnn},~ \varepsilon^{lf} \leq \Delta_{max}^{lf} &\to \textrm{use leap frog with posterior gradients for $N^{lf}$ samples}\\
    \varepsilon^{hnn} > \Delta_{max}^{hnn},~ \varepsilon^{lf} > \Delta_{max}^{lf} &\to \textrm{terminate tree building; move to next sample}\\
    \end{aligned}
\end{equation}
where $\varepsilon^{hnn}$ and $\varepsilon^{lf}$ are the integration errors [equation \eqref{eqn:NUTS_2}] when using leap frog with L-HNN and posterior gradients, respectively.

\section{Results}

\begin{figure}[h]
\begin{subfigure}{0.32\textwidth}
\centering  
\includegraphics[width=\textwidth]{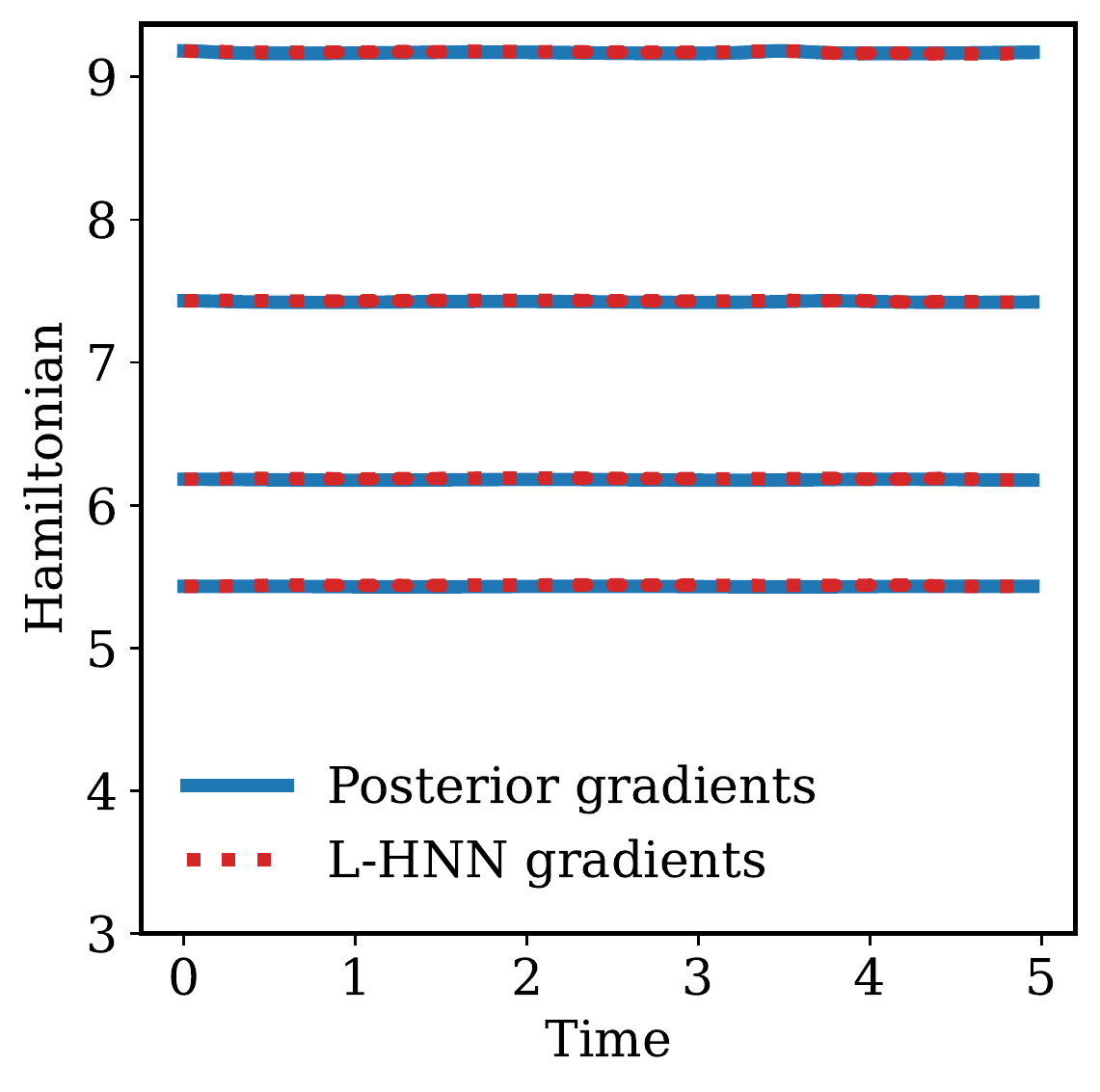} 
\caption{}
\label{2D_Gauss_1}
\end{subfigure}
\begin{subfigure}{0.32\textwidth}
\centering  
\includegraphics[width=\textwidth]{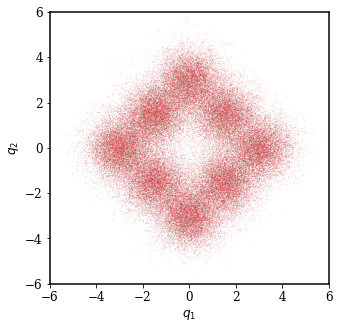} 
\caption{}
\label{2D_Gauss_2}
\end{subfigure}
\begin{subfigure}{0.32\textwidth}
\centering  
\includegraphics[width=\textwidth]{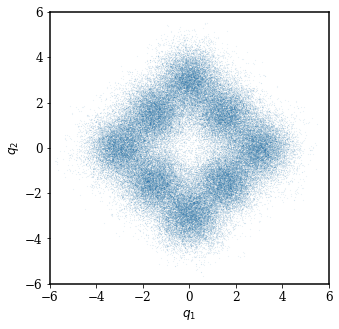} 
\caption{}
\label{2D_Gauss_3}
\end{subfigure}
\caption{A 2-D eight Gaussian mixture is considered as the target posterior. (a) Hamiltonian conservation for random initial $\{\pmb{q},\pmb{p}\}$ values considering L-HNN gradients and posterior gradients. (b) Samples from the target posterior simulated using LHNN-NUTS. (c) Samples from the target posterior simulated using traditional NUTS.}
\label{2D_Neal}
\end{figure}

We demonstrate Bayesian inference using L-HNNs in NUTS with online error monitoring considering several case studies. The performance is compared with traditional NUTS which relies on the posterior gradients. First, we consider a 2-D eight Gaussian mixture density. Each Gaussian has the same covariance which is an identity matrix but different mean vectors. Figure \ref{2D_Gauss_1} presents the Hamiltonian evolution with time for different initial $\{\pmb{q},\pmb{p}\}$ consider LHNN-NUTS and NUTS. It is noticed that LHNN-NUTS satisfactorily conserves the Hamiltonian values in a similar fashion to that of NUTS. Figures \ref{2D_Gauss_2} and \ref{2D_Gauss_3} present the scatter plot of $100,000$ simulated samples (first 5000 considered as burn-in) using LHNN-NUTS and NUTS, respectively. LHNN-NUTS captures all the posterior modes satisfactorily in a similar fashion to that of NUTS. Moreover, from a computational perspective, while NUTS requires more than 15 Million posterior gradients, LHNN-NUTS requires only 0.4 Million posterior gradients for the training data and during online error monitoring. The ESS per posterior gradient of LHNN-NUTS and NUTS is 0.0269 and 0.000707, respectively, which is between 1-2 orders of magnitude in improvement. 


We have also tested the performance of LHNN-NUTS considering several other complex high-dimensional posteriors. Table \ref{Table_Summary} summarizes the results in terms of the number of gradients and the ESS per posterior gradient and also presents these values corresponding to NUTS for comparison. Overall, it can be observed that LHNN-NUTS requires about 1-2 orders of magnitude less posterior gradients than NUTS. Moreover, LHNN-NUTS results in about an order of magnitude improvement in the ESS per posterior gradient compared to NUTS. These results contribute to the scalability and efficiency of the proposed LHNNs in NUTS with online error monitoring scheme relative to NUTS which is widely adopted for Bayesian inference.

\begin{table}[h]
\centering
\caption{Summary of the performance comparison between LHNN-NUTS and NUTS (adapted from Dhulipala et al. \cite{dhulipala2022b}).}
\label{Table_Summary}
\small
\begin{tabular}{ |c|c|c|c|c| }
\hline
\textbf{Posterior density} & \textbf{QOI} & \textbf{LHNN-NUTS} & \textbf{NUTS}\\
\hline
2-D eight Gaussian mixture & \Centerstack[c]{\# gradients\\ ESS/gradient} & \Centerstack[c]{0.4 Million\\ 0.0269} & \Centerstack[c]{15 Million\\ 0.000707}\\
\hline
10-D Rosenbrock density & \Centerstack[c]{\# gradients\\ ESS/gradient} & \Centerstack[c]{0.42 Million \\ 0.0368} & \Centerstack[c]{7 Million \\  0.00219}\\
\hline
24-D Bayesian logistic regression & \Centerstack[c]{\# gradients\\ ESS/gradient} & \Centerstack[c]{0.4 Million \\  0.243} & \Centerstack[c]{1.2 Million \\  0.0777}\\
\hline
100-D rough well & \Centerstack[c]{\# gradients\\ ESS/gradient} & \Centerstack[c]{0.42 Million \\  0.0065} & \Centerstack[c]{1.28 Million \\  0.00373}\\
\hline
\end{tabular}
\end{table}
\normalsize

\section{Summary and conclusions}

Physics-based neural networks that learn Hamiltonian systems (e.g., HNNs) have been receiving a great deal of interest. We proposed using HNNs to efficiently solve Bayesian inference problems by not requiring numerous posterior gradients. To increase the expressivity of HNNs and reduce the integration error while simulating Hamiltonian trajectories, we proposed HNNs with latent outputs (L-HNNs). We further proposed to use L-HNNs in NUTS with an online error monitoring scheme that reverts to using the posterior gradients for a few samples whenever the L-HNNs integration errors are high, which may be in regions of the uncertainty space where L-HNNs had little training data. Considering several complex high-dimensional posteriors, we demonstrated LHNN-NUTS with online error monitoring and compared its performance with NUTS. Overall, LHNN-NUTS with online error monitoring required 1-2 orders of magnitude lesser posterior gradients and resulted in about 1 order of magnitude better ESS per gradient compared to NUTS.



\section*{Acknowledgments}

This research is supported through INL's Laboratory Directed Research and Development (LDRD) Program under U.S. Department of Energy (DOE) Idaho Operations Office contract no. DE-AC07-05ID14517. 


\newpage

\section*{Broader impact}

Bayesian inference is a principal method for parameter calibration and uncertainty quantification in many scientific and engineering fields. Since closed-form solutions usually do not exist for practical Bayesian inference problems, Markov chain Monte Carlo (MCMC) algorithms are employed to sample from the target probability density. While random-walk MCMC algorithms are popular, they have poor scalability with the number of dimensions, and can feature large serial correlations between the samples. Therefore, algorithms under the HMC framework (including NUTS) are often employed. However, the HMC-based algorithms require numerous costly gradient estimations of the target posterior and can be expensive in practice. The proposed LHNN-NUTS with online error monitoring removes this computational hurdle and contributes to solving practical Bayesian inference problems across science and engineering while guaranteeing robustness.





\bibliographystyle{abbrv}
\bibliography{\jobname} 

\section*{Appendix A: Algorithms for L-HNNs in NUTS with online error monitoring}

\begin{algorithm}
\caption{L-HNNs in NUTS with online error monitoring (main loop) \cite{dhulipala2022b}}
\label{alg:HNN_NUTS1}
\begin{algorithmic}[1]
\State{Hamiltonian: $H = U(\pmb{q}) + K(\pmb{p})$, Samples: M; Starting sample: $\{\pmb{q}^0,~\pmb{p}^0\}$; Step size: $\Delta t$; Threshold for leapfrog: $\Delta_{max}^{lf}$; Threshold for L-HNNs: $\Delta_{max}^{hnn}$; Number of leapfrog samples: $N_{lf}$}
\State{Initialize $\pmb{1}_{lf} = 0,~n_{lf} = 0$}
\For{$i = 1:M$}
\State{$\pmb{p}(0) \sim \mathcal{N}(\pmb{0},~\pmb{I}_d)$}
\State{$\pmb{q}(0) = \pmb{q}^{i-1}$}
\State{$u \sim Uniform\Big(\Big[0,~\exp{\{-H\big(\pmb{q}(0),~\pmb{p}(0)\big)\}}\Big]\Big)$}
\State{Initialize $\pmb{q}^- = \pmb{q}(0),\pmb{q}^+ = \pmb{q}(0),\pmb{p}^- = \pmb{p}(0),\pmb{p}^+ = \pmb{p}(0),j=0,\pmb{q}^* = \pmb{q}^{i-1},n=1,s=1$}
\If{$\pmb{1}_{lf} = 1$}
\State{$n_{lf} \gets n_{lf} + 1$}
\EndIf
\If{$n_{lf} = N_{lf}$}
\State{$\pmb{1}_{lf} = 0,~n_{lf} = 0$}
\EndIf
\While{$s=1$}
\State{Choose direction $\nu_j \sim Uniform(\{-1,~1\})$}
\If{$j=-1$}
\State{$\pmb{q}^-,\pmb{p}^-,\_,\_,\pmb{q}^\prime,\pmb{p}^\prime,n^\prime,s^\prime,\pmb{1}_{lf} = \textrm{\textbf{BuildTree}}(\pmb{q}^-,\pmb{p}^-,u,\nu_j,j,\Delta t,\pmb{1}_{lf})$}
\Else
\State{$\_,\_,\pmb{q}^+,\pmb{p}^+,\pmb{q}^\prime,\pmb{p}^\prime,n^\prime,s^\prime,\pmb{1}_{lf} = \textrm{\textbf{BuildTree}}(\pmb{q}^+,\pmb{p}^+,u,\nu_j,j,\Delta t,\pmb{1}_{lf})$}
\EndIf
\If{$s^\prime = 1$}
\State{With probability $\textrm{min}\{1,~\frac{n^\prime}{n}\}$}, set $\{\pmb{q}^{i},\pmb{p}^{i}\} \gets \{\pmb{q}^\prime,\pmb{p}^\prime\}$ 
\EndIf
\State{$n \gets n + n^\prime$}
\State{$s \gets s^\prime \pmb{1}\big[(\pmb{q}^+ - \pmb{q}^-) \cdot \pmb{p}^- \geq 0\big] \pmb{1}\big[(\pmb{q}^+ - \pmb{q}^-) \cdot \pmb{p}^+ \geq 0\big]$}
\State{$j \gets j + 1$}
\EndWhile
\EndFor
\end{algorithmic}
\end{algorithm}
\normalsize

\begin{algorithm}
\caption{L-HNNs in NUTS with online error monitoring (build tree function) \cite{dhulipala2022b}}
\label{alg:HNN_NUTS2}
\begin{algorithmic}[1]
\State{function $\textrm{\textbf{BuildTree}}(\pmb{q},\pmb{p},u,\nu,j,\Delta t,\pmb{1}_{lf})$}
\If{$j=0$}
\State{Base case taking one leapfrog step}
\State{$\pmb{q}^\prime,\pmb{p}^\prime \gets$ Algorithm \ref{alg:HNN_eval} with initial conditions: $\pmb{z}(0)=\{\pmb{q},~\pmb{p}\}$, Steps: $1$; End Time: $\Delta t$}
\State{$\pmb{1}_{lf} \gets \pmb{1}_{lf}~\textrm{\textbf{or}}~\pmb{1}\big[H\big(\pmb{q}^\prime,~\pmb{p}^\prime\big)+\ln{u} > \Delta_{max}^{hnn}\big]$}
\State{$s^\prime \gets \pmb{1}\big[H\big(\pmb{q}^\prime,~\pmb{p}^\prime\big)+\ln{u} \leq \Delta_{max}^{hnn}\big]$}
\If{$\pmb{1}_{lf} = 1$}
\State{$\pmb{q}^\prime,\pmb{p}^\prime \gets$ Leapfrog integration with initial conditions: $\pmb{z}(0)=\{\pmb{q},~\pmb{p}\}$, Steps: $1$; End Time: $\Delta t$}
\State{$s^\prime \gets \pmb{1}\big[H\big(\pmb{q}^\prime,~\pmb{p}^\prime\big)+\ln{u} \leq \Delta_{max}^{lf}\big]$}
\EndIf
\State{$n^\prime \gets \pmb{1}\big[u \leq \exp{\{-H\big(\pmb{q}^\prime,~\pmb{p}^\prime\big)\}}\big]$}
\State{\textbf{return} $\pmb{q}^\prime,~\pmb{p}^\prime,\pmb{q}^\prime,~\pmb{p}^\prime,\pmb{q}^\prime,\pmb{p}^\prime,n^\prime,s^\prime,\pmb{1}_{lf}$}
\Else 
\State{Recursion to build left and right sub-trees (follows from Algorithm 3 in \cite{Hoffman2014}, with $\pmb{1}_{lf}$ additionally passed to and retrieved from every \textrm{\textbf{BuildTree}} evaluation)}
\State{\textbf{return} $\pmb{q}^-,~\pmb{p}^-,\pmb{q}^+,~\pmb{p}^+,\pmb{q}^\prime,\pmb{p}^\prime,n^\prime,s^\prime,\pmb{1}_{lf}$}
\EndIf
\end{algorithmic}
\end{algorithm}
\normalsize

\begin{algorithm}
\caption{Latent Hamiltonian neural networks evaluation in leapfrog integration \cite{dhulipala2022b}}
\label{alg:HNN_eval}
\begin{algorithmic}[1]
\State{Hamiltonian: $H$; Initial conditions: $\pmb{z}(0)=\{\pmb{q}(0),~\pmb{p}(0)\}$; Dimensions: $d$; Steps: $N$; End time: $T$}
    \State{$\Delta t = \frac{T}{N}$}
    \For{$j = 0:N-1$}
        \State{$t = j~\Delta t$}
        \State{Compute HNN output gradient $\frac{\partial H_{\pmb{\theta}}}{\partial \pmb{q}(t)}$}
        \For{$i = 1:d$}
            \State{$q_i(t+\Delta t) = q_i(t) + \frac{\Delta t}{m_i}~p_i(t) - \frac{\Delta t^2}{2m_i}~\frac{\partial H_{\pmb{\theta}}}{\partial q_i(t)}$}
        \EndFor
        \State{Compute HNN output gradient $\frac{\partial H_{\pmb{\theta}}}{\partial \pmb{q}(t+\Delta t)}$}
        \For{$i = 1:d$}
            \State{$p_i(t+\Delta t) = p_i(t) - \frac{\Delta t}{2}~\bigg(\frac{\partial H_{\pmb{\theta}}}{\partial q_i(t)} + \frac{\partial H_{\pmb{\theta}}}{\partial q_i(t+\Delta t)}\bigg)$}
        \EndFor
    \EndFor
\end{algorithmic}
\end{algorithm}
\normalsize

\newpage
\section*{Checklist}


\begin{enumerate}

\item For all authors...
\begin{enumerate}
  \item Do the main claims made in the abstract and introduction accurately reflect the paper's contributions and scope?
    \answerYes{}
  \item Did you describe the limitations of your work?
    \answerNo{}
  \item Did you discuss any potential negative societal impacts of your work?
    \answerNA{}
  \item Have you read the ethics review guidelines and ensured that your paper conforms to them?
    \answerYes{}
\end{enumerate}

\item If you are including theoretical results...
\begin{enumerate}
  \item Did you state the full set of assumptions of all theoretical results?
    \answerNA{}
        \item Did you include complete proofs of all theoretical results?
    \answerNA{}
\end{enumerate}

\item If you ran experiments...
\begin{enumerate}
  \item Did you include the code, data, and instructions needed to reproduce the main experimental results (either in the supplemental material or as a URL)?
    \answerNA{}
  \item Did you specify all the training details (e.g., data splits, hyperparameters, how they were chosen)?
    \answerNA{}
        \item Did you report error bars (e.g., with respect to the random seed after running experiments multiple times)?
    \answerNA{}
        \item Did you include the total amount of compute and the type of resources used (e.g., type of GPUs, internal cluster, or cloud provider)?
    \answerNA{}
\end{enumerate}

\item If you are using existing assets (e.g., code, data, models) or curating/releasing new assets...
\begin{enumerate}
  \item If your work uses existing assets, did you cite the creators?
    \answerNA{}
  \item Did you mention the license of the assets?
    \answerNA{}
  \item Did you include any new assets either in the supplemental material or as a URL?
    \answerNA{}
  \item Did you discuss whether and how consent was obtained from people whose data you're using/curating?
    \answerNA{}
  \item Did you discuss whether the data you are using/curating contains personally identifiable information or offensive content?
    \answerNA{}
\end{enumerate}

\item If you used crowdsourcing or conducted research with human subjects...
\begin{enumerate}
  \item Did you include the full text of instructions given to participants and screenshots, if applicable?
    \answerNA{}
  \item Did you describe any potential participant risks, with links to Institutional Review Board (IRB) approvals, if applicable?
    \answerNA{}
  \item Did you include the estimated hourly wage paid to participants and the total amount spent on participant compensation?
    \answerNA{}
\end{enumerate}

\end{enumerate}

\end{document}